\documentclass[10pt,journal,letterpaper,compsoc]{IEEEtran}
\usepackage[nocompress]{cite}
\usepackage[pdftex]{graphicx}
\usepackage[cmex10]{amsmath}
\usepackage{amssymb,amsthm}
\usepackage[utf8]{inputenc}
\usepackage[T1]{fontenc}
\theoremstyle{plain}
\newtheorem{theorem}{Theorem}
\newtheorem{lemma}{Lemma}
\newtheorem{definition}{Definition}

\theoremstyle{definition}
\newtheorem{example}{Example}
\newtheorem{algorithm}{Algorithm}

\theoremstyle{remark}

\newcommand{\SP}{\phantom{....}}

\DeclareMathOperator*{\argmin}{arg\hspace{.05em}min}
\DeclareMathOperator*{\argmind}{arg{\scriptstyle(d)}min}
\providecommand{\abs}[1]{\lvert#1\rvert}

\newcommand{\struct}[1]{\mathbb{#1}}
\newcommand{\decline}{\textit{''discard''}}
\newcommand{\Sol}{\mathrm{Sol}}
\begin{document}

\author{Michail~Schlesinger, 
  Boris~Flach, 
  and~Evgenij~Vodolazskiy%
  \IEEEcompsocitemizethanks{\IEEEcompsocthanksitem M. Schlesinger and E. 
Vodolazskiy are with International Research and Training Centre
of Information Technologies and Systems, National Academy of Science of Ukraine
\IEEEcompsocthanksitem B. Flach is with Czech Technical University in Prague, 
Czech Republic}}
\IEEEcompsoctitleabstractindextext{%
\begin{abstract}
The article considers  one of the possible generalizations of constraint 
satisfaction problems where relations are replaced by multivalued membership 
functions. In this case operations of disjunction and conjunction are replaced 
by maximum and minimum, and consistency of a solution becomes multivalued rather 
than binary. The article studies the problem of finding $d$ most admissible 
solutions for a given $d$. A tractable subclass of these problems is defined by 
the concepts of invariants and polymorphisms similar to the classic constraint 
satisfaction approach. These concepts are adapted in two ways. Firstly, the 
correspondence of "invariant-polymorphism" is generalized to (min,max) 
semirings. Secondly, we consider non-uniform polymorphisms, where each variable 
has its own operator, in contrast to the case of one operator common for all 
variables. The article describes an algorithm that finds $d$ most admissible 
solutions in polynomial time, provided that the problem is invariant with 
respect to some non-uniform majority operator. It is essential that this 
operator needs not to be known for the algorithm to work. Moreover, even a 
guarantee for the existence of such an operator is not necessary. The algorithm 
either finds the solution or discards the problem. The latter is possible only 
if the problem has no majority polymorphism.
\end{abstract}
\begin{IEEEkeywords}constraint satisfaction, discrete optimization, 
labeling, invariants, polymorphisms.\end{IEEEkeywords}
}

\title{M-best solutions for a class of fuzzy constraint satisfaction problems}

\maketitle

\IEEEdisplaynotcompsoctitleabstractindextext

\IEEEpeerreviewmaketitle

\section{\label{Introduction}Introduction.}
The constraint satisfaction problem (CSP) \cite{Rossi:HCP2006} is one of the 
paradigms of machine intelligence. The problem is to find values for variables  
satisfying a given set of constraints or to determine inconsistency of the 
constraints. The set of all possible constraint satisfaction problems forms an 
NP-complete class. However, three tractable subclasses are known. Each of these 
subclasses is defined in terms of polymorphisms 
\cite{Bulatov:SIAM2005,Cohen:HCP2006}, i.e.~operators under which the problem is 
invariant. The article considers sets of constraints invariant under majority 
operators.

A stronger version is the counting CSP, where the goal is to count the number of 
 solutions of a CSP rather than merely to decide if a solution exists. The 
complexity of counting CSPs has been analyzed in papers 
\cite{Bulatov:ACM2013,Bulatov:IC2007}. Evidently, this problem is stronger than 
the consistency problem, because any algorithm that solves the counting problem 
can be used to determine consistency. Unfortunately, the counting problem turns 
out to be much harder. The three known tractable subclasses of constraint 
satisfaction problems become NP-complete for the counting problem. Under some 
additional conditions only problems invariant under Maltsev operators are 
tractable \cite{Bulatov:ACM2013,Bulatov:IC2007}. This essential difference 
between consistency and counting problems makes it worthwhile to state and 
analyze intermediate problems. We are interested in a problem that is weaker 
than the counting problem but still stronger than the consistency problem. The 
problem is to determine whether a given set of constraints has more than $d$ 
solutions where $d$ is a given number. To the best of our knowledge this problem 
has not been stated yet, let alone analyzed.

One of our results is an algorithm which proves whether a given set of 
constraints  has more than $d$ solutions, provided that the constraints are 
invariant under a majority operator. The task is solved in polynomial time, 
avoiding an NP-complete counting problem. In case of a positive answer and 
$d>0$, the algorithm returns $d$ possible solutions for the given set of 
constraints. This particular result is closest to traditional constraint 
satisfaction theory. The article as a whole covers a more general set of 
questions.

We consider one of the possible modifications of constraint satisfaction 
problems with multilevel constraints. Instead of simply categorizing solutions 
into consistent and inconsistent ones, they rather define a level of 
consistency. This modification can be interpreted as fuzzy constraint 
satisfaction problem \cite{Dubois:CFS1993,Ruttkay:ICFS1994}, where the problem 
is to find the solution with highest level of consistency or, for the sake of 
brevity, the maximum admissible solution. The search of the maximum admissible 
solution can be reduced to discrete optimization tasks for special functions and 
proves to be tractable if the problem has a majority polymorphism 
\cite{schlesingerMoscow}. The main novelty of the present article is to show 
that $d$ best solutions (and not only the most admissible one) can be found in 
polynomial time under the same assumptions. The equivalent task in the context 
of standard constraint satisfaction problems is to determine whether the number 
of solutions is greater than $d$. The exact formulation of the result is given 
in Section \ref{Definitions} after the main definitions. Section \ref{Relations} 
explains relations to known results. 
\section{\label{Definitions} Problem definition and main result.}
The article uses the denotation $\argmind_{i \in I} f(i)$ similar to 
the commonly used denotation  $\argmin_{i \in I} f(i)$. 
\begin{definition} \label{MainDefinition}
  For a finite set $I$, an ordered set $W$, a function $f: I\rightarrow W$ and 
  an integer $0 < d < \abs{I}$, the expression $I^*=\argmind_{i \in I}f(i)$means 
  that $I^*$ is a subset of $I$ such that $\abs{I^*}=d$ and $f(i) \leqslant 
f(j)$ holds for any pair $i \in I^*$, $j \notin I^*$. For  $d \ge \abs{I}$ the 
  expression $I^*=\argmind_{i \in I}f(i)$ means that $I^*=I$.
\end{definition}
The subset $I^* \subset I$ specified by this definition is not 
necessarily unique, in the same way as the element $i^* \in I$ 
defined by the expression $i^* = \argmin_{i \in I}f(i)$ is not necessarily 
unique. The set $I^*$ is equivalently defined  by the inequalities 
\begin{equation} \label{MinMaxProperty}
\max_{i \in I^*} f(i)  \leqslant \max_{i \in I'} f(i), 
\end{equation}
which must be fulfilled for any subset $I' \subset I$ with $d$ elements.

Let $T$ and $K$ be two finite sets called the set of objects and the set of 
labels. A function $\bar{x}: T \rightarrow K$ will be called a labeling. Let 
$x_i$ denote the value of a labeling $\bar{x}:T \rightarrow K$  for an object 
$i \in T$ and let $x_S$ denote its restriction to a subset $S \subset T$. Let 
$K^S$ denote the set of all possible labellings $x_S: S \rightarrow K$ for any 
$S \subset T$ . Whenever we want to stress that the domain of a labeling 
is a union of a pair of disjoint subsets $A,B \subset T$, the labeling will be 
denoted by $(x_A, x_B)$, and not by $x_{A \cup B}$. Let $2^T$ denote the set of 
all possible subsets $S \subset T$. A set $\struct{S} \subset 2^T$ will be 
called a structure of the set $T$, the number $\max_{S \in 
\struct{S}}|S|$ being the order of a structure.

Let $W$ be a totally ordered set, $\struct{S}$ be a structure and let  
$\varphi_S:K^S \rightarrow W$ be a function given for each structure element $S 
\in \struct{S}$. We assume that each of these functions is defined by a table 
${\rm Tab}(S)=\left\{ \bigl(x, \varphi_S(x)\bigr) \mid x \in K^S \right\}$.
\begin{definition}
  The input data of a minimax labeling problem or, simply a problem, is a 
  quintuple
  \begin{equation} \label{MinMaxGeneral}
  \Phi = \bigl\langle T,K,W, \struct{S} \subset 2^T,(\varphi_S:K^S \rightarrow
  W \mid S \in \struct{S}) \bigr\rangle.
  \end{equation}
  The order of the problem $\Phi$ is defined as the order of the structure 
  $\struct{S}$.
\end{definition} 
The article considers arbitrary but fixed sets $K$ and $W$. Therefore, we refer 
to problems also in form of a triple $\Phi =\bigl\langle T, 
\struct{S},(\varphi_S \mid S \in \struct{S}) \bigr\rangle$ and not by a 
quintuple (\ref{MinMaxGeneral}).

The input data of a problem $\Phi$ define its objective function $\varphi:K^T 
\rightarrow W$ with values $\varphi(\bar{x})= \max_{S \in 
\struct{S}}\varphi_S(x_S)$, $\bar{x} \in K^T$, where $x_S$ is the restriction of 
$\bar{x}$ to $S$.
\begin{definition}
  For a given positive integer $d$ the solution of a problem $\Phi$ is a subset 
  $\Sol(\Phi)=\argmind\limits_{\bar{x} \in K^T}\varphi(\bar{x})$.
\end{definition}
The set of problems (\ref{MinMaxGeneral}) forms an NP-complete class, because 
any constraint satisfaction problem can be expressed in this format. We 
formulate a tractable subclass of such problems based on the concepts of 
polymorphisms and invariants, which are the main tools for tractability 
analysis of constraint satisfaction problems 
\cite{Bulatov:SIAM2005,Cohen:HCP2006}. We generalize these concepts in order to 
analyze problems (\ref{MinMaxGeneral}), which are more general than constraint 
satisfaction problems. 

Let $p_i: K \times K \times K \rightarrow K$ be a ternary operator defined for 
each $i \in T$. A collection $P=(p_i \mid i \in T)$ of such operators is 
understood as an operator $K^S \times K^S \times K^S \rightarrow K^S$, defined 
for each $ S \subset T$. Applying it to a triple $x,y,z \in K^S$ gives the 
labeling $P(x,y,z) = v \in K^S$ defined by $v_i=p_i(x_i,y_i,z_i)$, $i \in S$.
\begin{definition} \label{GroupPolInf}
  A function $\varphi_S \colon K^S \rightarrow W $ is invariant under the 
  operator $P=(p_i \mid i \in T)$ and an operator $P$ is a polymorphism of the 
  function $\varphi_S$ if the inequality $\max \bigl\{\varphi_S(x),   
  \varphi_S(y), \varphi_S(z)\bigr\} \geqslant \varphi_S\bigl(P(x,y,z)\bigr)$  
  holds for each triple $x,y,z \in K^S$.
\end{definition}
This definition was first introduced in \cite{Bulatov:LNCS2003} and is 
more general than the polymorphism-invariant correspondence commonly used in 
constraint satisfaction theory. Definition~\ref{GroupPolInf} assumes 
that each object $i\in T$ gets assigned its own operator $p_i: K \times K 
\times K \rightarrow K$, instead of assigning a single operator to all 
variables. If it is necessary to emphasize that the components $p_i$ of an 
operator $P=(p_i \mid i \in T)$ depend on $i$, and, may differ from each 
other, we call the operator non-uniform. 
\begin{definition} \label{ProblemPolInf}
  An operator $P=(p_i \mid i \in T)$ is a polymorphism of the problem 
  $\Phi=\bigl\langle T, \struct{S},(\varphi_S \mid S \in \struct{S}) 
  \bigr\rangle$, and a problem $\Phi$ is invariant under the operator $P$ if 
  $P$   is a polymorphism of all functions $\varphi_S$, $S \in 
\struct{S}$.
\end{definition}
\begin{definition} \label{Majority1}
  An operator $P=(p_i \mid i \in T)$ is a majority operator if the equalities
  \begin{equation*}
  p_i(y, x, x)=p_i(x, y, x)=p_i(x, x, y)=x
  \end{equation*}
  hold for all $i \in T$ and for all $x, y\in K$.
\end{definition}
The result of this paper is an algorithm that solves problems 
(\ref{MinMaxGeneral}) if they have a majority polymorphism. Its time complexity 
depends on parameters $\abs{T}$, $\abs{K}$, $|\struct{S}|$, the number of 
required labellings $d$ and the total size $\sum_{S \in \struct{S}}|{\rm 
Tab}(S)|$ of the tables, which represent the functions $\varphi_S$, $S \in 
\struct{S}$. The main idea is to transform a problem of arbitrary order into an 
equivalent problem of order 2, and then to solve the second order problem by 
sequentially excluding variables. The order reduction procedure is described in 
Section \ref{OrderNto2}, the approach for solving second order 
problems is described in Section \ref{StarToSimplex}.

The set of problems (\ref{MinMaxGeneral}) solvable by the algorithm for $d=1$ 
includes a well known subclass of constraint satisfaction problems and its fuzzy 
modifications. Example~\ref{Cluster} illustrates the likely less known fact that 
certain clustering problems can be expressed in the form (\ref{MinMaxGeneral}). 
Example~\ref{Relaxation} shows, how solving (\ref{MinMaxGeneral}) for $d > 1$ 
can improve a certain workaround for solving problems with additional global 
constraints. 
\begin{example} {\bfseries Clustering. \label{Cluster}} 
Consider a finite set $T$ and a function $r:T \times T \rightarrow W$ defining a 
dissimilarity $r(s,t)$ for each pair $s,t \in T$. A partition of the set $T$ 
into two subsets is a pair $(T_1,T_2)$ such that  $T_1 \cup T_2 = T$, $T_1 \cap 
T_2 = \emptyset$ and its quality is defined by the value
\begin{equation*} \nonumber 
 F(T_1,T_2) = \max \Bigl\{ 
 \max\limits_{s, t \in T_1}r(s,t),
 \max\limits_{s, t \in T_2}r(s,t) \Bigr\}.
\end{equation*}
One possible definition of a clustering problem is to find the best partition
\begin{equation} \label{clustering}
 (T_1^*,T_2^*)=\argmin_{(T_1,T_2)}\max \Bigl\{
 \max_{s, t \in T_1}r(s,t),
 \max_{s, t \in T_2}r(s,t) \Bigr\} .
\end{equation}
This problem is reduced to a minimax problem (\ref{MinMaxGeneral}) by
\begin{gather*} \nonumber
 K=\{1,2\}, \quad \struct{S} = 
 \bigl\{ \{s,t\} \bigm | s,t \in T, t \ne s \bigr\}, \\
 \varphi_{\{s,t\} }(k,k') = 
 \begin{cases}
    r(s,t) &\text {if $k=k'$,} \\
    \min W &\text {otherwise}
 \end{cases}, \hspace{.5em}
 \{ s,t\} \in \struct{S}.
\end{gather*}
A solution $\bar{x}^* = \argmin_{\bar{x} \in K^T} \max_{ \{ s,t \} 
\in \struct{S} }\varphi_{\{s,t\} }(x_{s},x_{t})$ of this labeling problem 
defines a solution of the clustering problem (\ref{clustering}) via $T_k^* = 
\{s \in T \mid x_s^* = k  \}, k \in \{ 1,2 \}$.

The minimax problem has a binary label domain $K$ and is of order two. Any 
such problem is invariant under some majority operator and can be solved by the 
provided algorithm.
\hfil\qed\end{example}
\begin{example} {\bfseries Constraint relaxation. \label{Relaxation}}
Suppose that the task is to find the best labeling for given data 
(\ref{MinMaxGeneral}) under some additional constraints. Formally put, the 
labeling must belong to some given set $\cal{K}$ of labellings. This might be 
a condition which is easy to verify. For example, it might be required that a 
certain label $k_0 \in K$ appears in a labeling $\bar{x}$ at most $l$ times.
However, seeking the best labeling
\begin{equation} \label{Restrict}
\bar{x}^* = \argmin_{\bar{x} \in \cal{K} }\bigl[ \max_{S \in \struct{S}}
\varphi_S( x_S) \bigr] 
\end{equation}
under such additional constraints may turn out to be much harder than seeking 
the best labeling 
\begin{equation} \label{Relaxed}
\bar{x}^* = \argmin_{\bar{x} \in {K^T}}\bigl[ \max_{S \in \struct{S}}
\varphi_S (x_S) \bigr].
\end{equation}
without such constraints. Moreover, it might happen that the additional 
constraints are hard to formalize. The set $\cal{K}$ may represent, for example, 
a user who rejects labellings based on informal personal preferences.

A workaround is to find the best labeling (\ref{Relaxed}) and to check condition 
$\bar{x}^* \in \cal{K}$ afterwards. Obviously, if the condition holds, then 
$\bar{x}^*$ is a solution of (\ref{Restrict}). However, this requirement is 
rather too strong.  It can be weakened by finding $d$ best labellings. The 
approach for solving (\ref{Restrict}) is to consider labellings one by one, from 
best to worst. The first labeling in the sequence which fulfills $\bar{x}^* \in 
\cal{K}$ is a solution of the task $\argmin_{\bar{x} \in \cal{K} }\bigl[ \max_{ 
S \in \struct{S}} \varphi_S( x_S) \bigr]$. Of course, this labeling may appear 
late in the sequence and the problem (\ref{Restrict}) will remain unsolved. 
However, an incorrect solution is excluded in any case.
\qed\end{example}
\section{\label{Relations} Relations to known results.}
The closest counterpart to problem~(\ref{MinMaxGeneral}) are constraint 
satisfaction problems. It is known that constraint satisfaction problems with a 
majority polymorphism form a tractable subclass \cite{Jeavons:AI1998}. This 
result can be easily generalized to problems~(\ref{MinMaxGeneral}) for $d=1$. 
Solving $\argmin_{\bar{x} \in K^T}\varphi(\bar{x})$ is tractable because it can 
be reduced to solving $\log(\abs{K} \times \abs{T})$ constraint satisfaction 
problems.

We solve the task for arbitrary $d$. For constraint satisfaction problems this 
means to prove existence of at least $d$ solutions satisfying the constraints. 
As far as we know, this question has not yet been studied for constraint 
satisfaction problems.

The article presents an algorithm that solves a certain subclass of an 
$NP$-complete class of problems. This subclass is defined in terms of existence 
of a non-uniform majority polymorphism. For practical application of the 
algorithm, it is necessary to either know its behavior on problems instances 
without such a polymorphism or to have a method for proving existence of a 
non-uniform majority polymorphism for a given problem instance. There are known 
methods for proving whether a problem has a uniform majority polymorphism 
\cite{Cohen:HCP2006}, however, we are not aware of such a method for non-uniform 
majority polymorphisms. We conjecture this to be a nontrivial task. The 
advantage of the presented algorithm is that such a prior control of input data 
is not required. For any problem~(\ref{MinMaxGeneral}) from an $NP$-complete 
class given on its input, the algorithm stops in polynomial time either 
returning a set of $d$ best labellings or discarding the problem. The latter is 
possible only if the problem has no majority polymorphism. Therefore, the 
algorithm solves any problem (\ref{MinMaxGeneral}) that is invariant under some 
non-uniform majority operator and avoids to answer the potentially hard question 
of existence of such an operator let alone to find it.
\section{Transforming problems of arbitrary order to problems of second order. 
\label{OrderNto2}}
Let $T$ be a set of objects, $S \subset T$ and  $R= T \setminus S$.
\begin{definition} \label{Projection}
The projection of a function $\varphi: K^T \rightarrow W$ onto the subset $S$ 
is the function $\varphi_S:K^S \rightarrow W$, obtained by minimizing over all 
variables not in $S$, i.e.
\begin{equation*}
 \varphi_S(x_S)=\min_{x_R\in K^R}\varphi(x_S,x_R) .
\end{equation*}
\end{definition}
The following property immediately follows from this definition.
\begin{lemma}\label{ProjectOfProject}
Let $S \subset R \subset T$, $\varphi_S$ and $\varphi_R$ be the projections of a 
function $\varphi: K^T \rightarrow W$ onto $S$ and $R$ respectively and 
$\varphi_*$ be the projection of $\varphi_{R}$ onto $S$. Then $\varphi_* = 
\varphi_S$.
\end{lemma}
The next two lemmas express properties of functions invariant under some 
operator.
\begin{lemma}\label{PolProject}
If a function $\varphi: K^T \rightarrow W$ is invariant under an operator 
$P=(p_i \mid i \in T)$, then its projection $\varphi_S$ is invariant under the 
same operator.
\end{lemma}
\begin{IEEEproof}
Let us denote $R = T \setminus S$. Let $x_S$, $y_S$, $z_S$ be three labellings 
of the form $S \rightarrow K$. Since $\varphi_S$ is the projection of $\varphi$ 
onto $S$, there exist three labellings
\begin{equation*}
 \bar{x}=(x_S,x_R), \quad 
\bar{y}=(y_S,y_R),\quad \bar{z}=(z_S,z_R)
\end{equation*}
of the form $T \rightarrow K$, such that
\begin{equation*}
\varphi(\bar{x})=\varphi_S(x_S), \quad 
\varphi(\bar{y})=\varphi_S(y_S), \quad
\varphi(\bar{z})=\varphi_S(z_S).
\end{equation*}
Let us denote
\begin {equation*}
\bar{u} =P(\bar{x},\bar{y},\bar{z}),\; u_S =P(x_S,y_S,z_S), \; 
u_R=P(x_R,y_R,z_R).
\end{equation*}
Because $\varphi_S$ is the projection of $\varphi$ onto $S$ and $\varphi$ is 
invariant under $P$, it follows that
\begin{multline*}
 \varphi_S\bigl(P(x_S,y_S,z_S)\bigr) = \varphi_S(u_S) = 
 \min_{x_R\in K^R} \varphi(u_S,x_R) \leqslant \\
 \leqslant \varphi(u_S,u_R) = \varphi(\bar{u}) \leqslant 
 \max \bigl\{\varphi(\bar{x}),\varphi(\bar{y}), \varphi(\bar{z}) \bigr\} = \\
  = \max \bigl\{ \varphi_S(x_S),\varphi_S(y_S), \varphi_S(z_S) \bigr\}. 
  \qedhere 
\end{multline*}
\end{IEEEproof}
\begin{lemma}\label{PolMax}
If two functions $\varphi,\psi \colon K^T \rightarrow W$ are invariant under an 
operator $P=(p_i \mid i \in T)$, then their element-wise maximum, 
i.e.~the function $\omega:K^T \rightarrow W$ with values 
$\omega(\bar{x}) = \max \bigl\{\varphi(\bar{x}), \psi(\bar{x}) \bigr\}$, 
$\bar{x} \in K^T$, is invariant under the same operator.
\end{lemma}
\begin{IEEEproof}
Let $\bar{x}_i$, $i=1,2,3$ be three labellings and 
$\bar{y} = P(\bar{x}_1,\bar{x}_2, \bar{x}_3)$. 
The fact that $\varphi$ and $\psi$ are invariant under $P$ means that
$\max_i \varphi(\bar{x}_i) \geqslant \varphi(\bar{y})$ and 
$\max_i \psi(\bar{x}_i) \geqslant \psi(\bar{y})$. 
It follows that
\begin{equation*}
  \max_i
  \max\bigl\{\varphi(\bar{x}_i), \psi(\bar{x}_i)\bigr\} \geqslant
  \max \bigl\{ \varphi(\bar{y}), \psi(\bar{y})\bigr\},
\end{equation*}
and, equivalently,
\begin{equation*}
 \max_i \omega(\bar{x}_i) \geqslant 
 \omega(\bar{y}) = \omega\bigl(P(\bar{x}_1, \bar{x}_2, \bar{x}_3)\bigr).
\end{equation*} 
\end{IEEEproof}
If a function $\varphi:K^T\rightarrow W$ has a majority operator then it has an 
important additional property.
\begin{lemma}\label{Decomposition}
Let $\varphi \colon K^T \rightarrow W$ be a function which has a majority 
polymorphism and let $Q$, $R$, $S$ be pairwise disjoint subsets of $T$ such 
that $Q~\cup~R~\cup~S = T$. Denote by  $\varphi_{QR}$, $\varphi_{QS}$, 
$\varphi_{RS}$ the projections of $\varphi$ onto the subsets $Q \cup 
R$, $Q \cup S$, $R \cup S$ respectively. Then the equality
\begin{equation*}
\varphi(\bar{x}) = \max \bigl\{ 
\varphi_{QR}(x_Q,x_R),\; \varphi_{QS}(x_Q,x_S),\;
\varphi_{RS}(x_R,x_S) \bigr\}
\end{equation*}
holds for any labeling $\bar{x} = (x_Q, x_R, x_S) \in K^T$.
\end{lemma}
\begin{IEEEproof}
Let us pick an arbitrary labeling $\bar{x}$ for the following considerations. 
Let $x_Q$, $x_R$, $x_S$ denote the restrictions of the labeling $\bar{x}$ onto 
the subsets $Q$, $R$, $S$ respectively. By definition of projection the 
inequalities
\begin{align*}
&\varphi(\bar{x}) \geqslant \varphi_{QR}(x_Q,x_R),\\ 
&\varphi(\bar{x}) \geqslant \varphi_{QS}(x_Q,x_S),\\ 
&\varphi(\bar{x}) \geqslant \varphi_{RS}(x_R,x_S), 
\end{align*}
are valid, and, consequently we have
\begin{equation*}
\varphi(\bar{x})\geqslant \max \bigl\{ \varphi_{QR}(x_Q,x_R),\;
\varphi_{QS}(x_Q,x_S),\; \varphi_{RS}(x_R,x_S) \bigr\} . 
\end{equation*}
Let us prove the converse inequality
\begin{equation*}
\varphi(\bar{x}) \leqslant \max \bigl\{ \varphi_{QR}(x_Q,x_R),\;
\varphi_{QS}(x_Q,x_S),\; \varphi_{RS}(x_R,x_S) \bigr\}. 
\end{equation*}
Because $\varphi_{QR}$, $\varphi_{QS}$, $\varphi_{RS}$ are 
the projections of the function
$\varphi$ onto the subsets $Q \cup R$, $Q \cup S$, $R \cup S$, there exist 
three labellings
$y_S \colon S \rightarrow K$, $y_R \colon R \rightarrow K$ and 
$y_Q \colon Q\rightarrow K$ such that
\begin{align} \label{expansions}
&\varphi(x_Q, x_R, y_S) = \varphi_{QR}(x_Q, x_R),\nonumber \\ 
&\varphi(x_Q, y_R, x_S) = \varphi_{QS}(x_Q, x_S), \\
&\varphi(y_Q, x_R, x_S) = \varphi_{RS}(x_R, x_S). \nonumber
\end{align}
The function $\varphi$ has some majority polymorphism $P$, therefore, 
(\ref{expansions}) implies the chain
\begin{multline*}
 \varphi(\bar{x}) = \varphi(x_Q, x_R, x_S) = \\
 =\varphi\bigl(
 P(x_Q,x_Q,y_Q),
 P(x_R,y_R, x_R),P(y_S,x_S,x_S)\bigr) \leqslant \\
 \leqslant \max\bigl\{ 
 \varphi(x_Q, x_R, y_S), \varphi(x_Q, y_R, x_S),
 \varphi(y_Q, x_R, x_S)  \bigr\} = \\
 = \max\bigl\{ \varphi_{QR}(x_Q, x_R), \varphi_{QS}(x_Q, x_S),
 \varphi_{RS}(x_R, x_S)  \bigr\}. \qedhere
\end{multline*}
\end{IEEEproof}
Lemma~\ref{Decomposition} shows that any function $\varphi \colon K^T 
\rightarrow W$ of $\abs{T}$ arguments that has a majority polymorphism, can be 
expressed in terms of three projections $\varphi_A$, $\varphi_B$, $\varphi_C$ 
onto subsets $A,B,C \subset T$, provided that the union of their pairwise 
intersections coincides with $T$. Each of these functions depends on less 
variables than $\varphi$ and, according to Lemma~\ref{PolProject}, they are 
invariant under the same operator as $\varphi$. Therefore, each of the functions 
$\varphi_A$, $\varphi_B$, $\varphi_C$  can in turn be expressed in terms of 
functions of less variables. Moreover, there will be no collisions when 
projecting some functions $\varphi_A$ and $\varphi_B$ onto $D \subset A \cap B$. 
According to Lemma~\ref{ProjectOfProject}, both projections are equal to the 
projection of $\varphi$ onto $D$. Therefore, any function $\varphi \colon K^T 
\rightarrow W$ that has a majority polymorphism can be expressed in  form of 
$\varphi(\bar{x}) = \max_{i,j \in T} \varphi_{ij}(x_i,x_j)$, $\bar{x} \in K^T$, 
where $\varphi_{ij}$ are the projections of $\varphi$ onto $\{i,j\}$ and have 
the same majority polymorphism as $\varphi$.

The stated properties allow us to transform any problem with a majority 
polymorphism into an equivalent problem of second order, even if the 
polymorphism itself is not known. Instead of denoting the second order problem 
by a triple $\bigl\langle T,\struct{S},(\varphi_S \mid S \in 
\struct{S})\bigl\rangle$, we will denote it by a tuple $\bigl\langle 
T,(\varphi_{ij} \mid i, j \in T) \bigr\rangle$, where $\varphi_{ij}$, $i,j \in 
T$, are functions $K \times K \rightarrow W$ such that 
$\varphi_{ij}(k,k')=\varphi_{ji}(k',k)$ for all $i,j \in T$, $k,k' \in K$ and 
$\varphi_{ii}(k,k') = \min W$ for all $i \in T$. The value of the objective 
function for a labeling $\bar{x} \in K^T$ is $\max_{i,j \in T} 
\varphi_{ij}(x_i,x_j)$ 
\begin{theorem} Any problem $\Phi = \bigl\langle T, 
\struct{S},(\varphi_S \mid S \in \struct{S})\bigr\rangle$ that has a  
majority polymorphism can be transformed to a second order problem 
$\Psi = \bigl\langle T,(\psi_{ij} \mid i, j \in T) \bigr\rangle$ such that
\begin{equation*}
\max_{S \in \struct{S}} \varphi_S(x_S) = 
\max_{i,j \in T} \psi_{ij}(x_i,x_j), 
\quad \bar{x} \in K^T,
\end{equation*}
where $x_S$ is the restriction of labeling $\bar{x}$ onto $S \in \struct{S}$ 
and $x_i$ are its values for $i \in T$. The second order problem $\Psi$ is 
invariant under the same majority operator as $\Phi$.
\end{theorem}
\begin{IEEEproof}
Let us denote the projection of $\varphi_S$ onto $\{i,j\}\subset 
S$ by $\varphi_{ij}^S$. We assume $\varphi_{ij}^S(x_i,x_j)= \min W$ for 
$\{i,j\} \not\subset S$. The functions $\varphi_{ij}^S$ are invariant under the 
same operator as the function $\varphi_S$. Let us define functions $\psi_{ij}$ 
of the problem $\Psi$ as $\psi_{ij}=\max_{S \in\struct{S}}\varphi_{ij}^S$. 
According to  Lemma~\ref{PolMax} these functions are 
invariant under the same operator as $\varphi_{ij}^S$. Therefore, both problems 
$\Phi$ and $\Psi$ are invariant under the same majority operator. And, the 
following chain 
\begin{multline*}
 \max_{S \in \struct{S}} \varphi_S(x_S)
 =\max_{S \in \struct{S}}\max_{i,j \in S} \varphi_{ij}^S(x_i,x_j) = \\
 =\max_{i,j \in T}\max_{S \in \struct{S}}\varphi_{ij}^S(x_i,x_j)=
 \max_{i,j \in T} \psi_{ij}(x_i,x_j)
\end{multline*}
holds for each labeling $\bar{x} \in K^T$.
\end{IEEEproof}
The proof of the theorem implicitly contains an algorithm for transforming the 
problem $\Phi = \bigl\langle T, \struct{S},(\varphi_S \mid S \in \struct{S}) 
\bigr\rangle$ into the problem $\Psi = \bigl\langle T,(\psi_{ij} \mid i,j \in T) 
\bigr\rangle$. Assuming that the functions $\varphi_S$, ${\rm S}\in \struct{S}$, 
which define the problem $\Phi$, are given in form of a tables ${\rm 
Tab(S)}=\bigl\{(x, \varphi_S(x)) \mid x \in K^S \bigr\}$, this algorithm reads 
as follows. 
\begin{algorithm}{\bf Reducing the problem's order.}\label{ZnIntoZ2Complete}  \\
{\bf Input:} problem $\Phi = \bigl\langle T, \struct{S},(\varphi_S \mid S 
\in\struct{S}) \bigr\rangle$.\\
{\bf Output:} problem $\Psi = \bigl\langle T,(\psi_{ij} \mid i, j \in 
T)\bigr\rangle$. \\
\SP 0. For each $i, j \in T$, $k, k' \in K$ \\
\SP \SP \SP $\psi_{ij}(k,k')= \min W$;\\
\SP 1. for each $S \in \struct{S}$\\
\SP \SP 
\SP 1.0. for each $i, j \in S$, $k, k' \in K$ \\
\SP \SP \phantom{\{}\SP \SP\SP $\varphi_{ij}^S(k,k')= \max W$;\\
\SP \SP \phantom{\{}\SP 1.1. for each $ (\bar{x}, w) \in {\rm Tab(S)}$ and each
$i,j \in S$ \\
\SP \SP \phantom{\{}\SP \SP \SP $\varphi_{ij}^S(x_i,x_j)= \min
\bigl\{\varphi_{ij}^S(x_i,x_j), w\bigr\}; $\\
\SP \SP \phantom{\{}\SP 1.2. for each $ (\bar{x}, w) \in {\rm Tab(S)}$\\
\SP \SP \phantom{\{}\SP \SP \SP if $w \ne \max\limits_{i,j \in S} 
\varphi_{ij}^S(x_i,x_j)$,\\
\SP \SP \phantom{\{}\SP \SP \SP then return \decline; \\
\SP \SP \phantom{\{}\SP 1.3. for each  $i, j \in S$, $k, k' \in K$ \\
\SP \SP \phantom{\{}\SP \SP \SP $\psi_{ij}(k,k') = \max \bigl\{ \psi_{ij}(k,k'),
\varphi_{ij}^S(k,k')  \bigr\} $.
\qed
\end{algorithm}
If the input problem $\Phi$ has a majority polymorphism then 
Algorithm~\ref{ZnIntoZ2Complete} is  guaranteed to transform the problem into an 
equivalent problem $\Psi$ of order two. Testing conditions in p.1.2 is redundant 
in this case. None of them holds. However, testing these conditions extends the 
scope of the algorithm to cover any problem, and not only those which have a 
majority polymorphism. The absence of a {\decline} message guarantees that the 
algorithm has successfully converted the input problem into a second order 
problem. This is true regardless of presence or absence of a majority 
polymorphism. Notice however, that the resulting problem has no majority 
polymorphism if the input problem lacks one. As will be shown in 
section~\ref{StarToSimplex}, this does not violate the applicability of 
algorithms given there for solving problems of second order. The {\decline} 
message means that the problem $\Phi$ is not in the applicability range of the 
algorithm. This is possible only if the problem has no majority polymorphism.

The complexity of Algorithm~\ref{ZnIntoZ2Complete} as well as of all other 
presented algorithms is measured by the number of $\max \{w,w'\}$ and 
$\min\{w,w'\}$  operations. The complexity of Algorithm \ref{ZnIntoZ2Complete} 
depends polynomially on the parameters of the problem: the numbers $\abs{T}$, 
$\abs{K}$, $\abs{\struct{S}}$, the order $n$ and the size
$l=\sum_{S \in \struct{S}}\abs{{\rm Tab}(S)}$ of the input data.

The complexity of p.0 is of order $\abs{T}^2 \times \abs{K}^2$. The total 
complexity for all $S \in \struct{S}$ of p.1.0 and 1.3 is of order 
$\abs{\struct{S}} \times \abs{K}^2 \times n^2$. The total complexity for all $S 
\in \struct{S}$ of p.1.1 and 1.2 is of order $l \times n^2 $.
\section{\label{StarToSimplex}Second order problems.}
\subsection{\label{GeneralFrame}A general approach for excluding variables.}
Let us define two problems
 $\Phi = \bigl\langle T, (\varphi_{ij} \mid i, j \in T) \bigr\rangle$ 
 and
 $\Phi^* = \bigl\langle S, (\varphi_{ij}\mid i, j \in S )\bigr\rangle $,
for a set $T$ and the set $S = T\setminus \{ t \}$ obtained from $T$ by removing 
an arbitrary element $t\in T$. Consider an algorithm that ``reduces'' the search 
of a solution $\Sol(\Phi)$ to the search of $\Sol(\Phi^*)$ in a greedy way. 
Algorithms of this type minimize a function of $n$ variables by minimizing an 
auxiliary function of $(n-1)$ variables first, and then find the optimal value 
of the remaining $n$-th variable by keeping the other variables fixed to the 
previously found minimizer of the auxiliary function. The auxiliary function 
itself is minimized by the same greedy algorithm, so that the optimization over 
$n$ variables is eventually ``reduced'' to $n$ optimizations over a single 
variable.

We modify this idea in two ways. Firstly, a greedy algorithm is used to find $d$ 
best solutions, and not only the best solution. Secondly, we include an 
essential test that allows to detect situations in which the result of the 
algorithm differs from $\Sol(\Phi)$.

\begin{algorithm}{\bf Greedy algorithm.}\label{Trivial}  \\
{\bf Input:} a problem $\Phi = \bigl\langle T, (\varphi_{ij} \mid i, j \in T )
\bigr\rangle$. \\
{\bf Output:} $\Sol(\Phi)$ or a message \decline.\\ 
0. If $\abs{T}=2$, $T = \{ a,b \}$ \\
\phantom{0. } then $\Sol(\Phi) =
\argmind\limits_{(x_a, x_b) \in K^2 } \varphi_{ab}(x_a,x_b)$;\\
else\\
1. pick $t \in T$, let $S = T \setminus \{t\}$;\\
2. using Algorithm~\ref{Trivial} find
$\Sol\bigl(\Phi^*\bigr)$, \\
\SP\SP\SP $\Phi^* = \bigl\langle S, (\varphi_{ij} \mid i, j \in S 
)\bigr\rangle$;\\
3. if there is at least one labeling $x \in \Sol\bigl(\Phi^*\bigr)$ \\
\SP\SP\SP fulfilling the inequality\\
\SP\SP\SP $ \max\limits_{i,j \in S} \varphi_{ij}(x_i, x_j) < 
\min\limits_{k \in K } \max\limits_{i \in S} \varphi_{it}(x_i, k)$, \\ 
\phantom{3. } then return \decline;\\
4. construct the auxiliary set \\
\phantom{4. } $WORK = \bigl\{(x, x_t)  \in K^T \bigm| x \in \Sol(\Phi^*), x_t 
\in K 
\bigr\}$;\\
5.  find 
$\Sol' = \argmind\limits_{\bar{x} \in WORK}
\max\limits_{i,j \in T}\varphi_{ij}(x_i,x_j)$.
\hfil\qed\end{algorithm}
The subset $\Sol'$ returned by the algorithm is not necessarily the solution of 
the problem. However, testing conditions in p.3 allows to detect situations in 
which $\Sol'$ is a solution. The next lemma proves that $\Sol'$ is the required 
labeling subset if the algorithm does not output \decline.
\begin{lemma}\label{GreedySecurity} \phantom{}
  Let $\Phi = \bigl\langle T,(\varphi_{ij} \mid i, j \in T) \bigr\rangle$ and 
  $\Phi^* = \bigl\langle S,(\varphi_{ij} \mid i, j \in S ) \bigr\rangle $ be 
  two   problems such that $S=T\setminus \{t\}$ and $t \in T$. Let 
  \begin{equation*}
  WORK = \bigl\{(x, x_t)  \in K^T \bigm| x \in \Sol(\Phi^*), 
  x_t \in K  
  \bigr\}
  \end{equation*}
denote all possible extensions of labellings $x\in \Sol(\Phi^*)$ and let
\begin{equation*}
 \Sol'= \argmind\limits_{\bar{x} \in WORK} 
 \max\limits_{i,j\in T} 
 \varphi_{ij}(x_i,x_j).
\end{equation*}
be a set of $d$ best labellings in $WORK$. If  the inequality
\begin{equation} \label{IfLemma}
 \max_{i,j \in S} \varphi_{ij}(x_i,x_j) \geqslant
 \min_{k \in K } \max_{i \in S} \varphi_{ti}(k,x_i)
\end{equation} 
holds for each labeling $x \in \Sol(\Phi^*)$, then $\Sol'$ is a 
solution of $\Phi$, i.e.
$\Sol' =
 \argmind\limits_{\bar{ x} \in K^T} \max\limits_{i,j \in T}
 \varphi_{ij}(x_i,x_j) .$
\end{lemma} 
\begin{IEEEproof} Denote
\begin{align*}
 \varphi_S(x) &= \max\limits_{i,j \in S}\varphi_{ij}(x_i,x_j),
 &\theta^* &= \max \bigl\{\varphi_S(x) \mid x \in\Sol(\Phi^*)\bigr\}, \\
 \varphi(\bar{x}) &= \max\limits_{i,j \in T}\varphi_{ij}(x_i,x_j),
 &\theta' &= \max \bigl\{\varphi(\bar{x})\mid \bar{x} \in\Sol'\bigr\}
\end{align*}
for $x \in K^S$ and $\bar{x} \in \Sol'$.

Let us first prove that $ \theta' \leqslant \theta^*$. We enumerate labellings 
from $\Sol(\Phi^*)$ by numbers $l \in \{ 1, 2, \dots , d \}$ and denote by $x(l) 
\in \Sol(\Phi^*)$ the labeling with number $l$, so that $\Sol(\Phi^*)=\{x(l) 
\mid l=1,2, \dots , d\}$. Let us define the label $x_t(l) = \argmin_{k \in K} 
\max_{i \in S} \varphi_{it}(x_{i}(l), k)$ for each labeling $x(l) \in 
\Sol(\Phi^*)$. This gives $d$ labellings $\bigl(x(l), x_t(l) \bigr) \in WORK$, 
$l \in \{ 1, 2, \dots , d \}$. Because of assumption (\ref{IfLemma}) they 
fulfill the equalities
\begin{equation} \label{Just}
\varphi\bigl(x(l), x_t(l)\bigr) = \varphi_S\bigl(x(l)\bigr), \quad l \in \{
1,2,\ldots,d \},
\end{equation}
which leads to the chain
\begin{multline*} \nonumber 
\theta'=\max_{\bar {x} \in \Sol'} \varphi(\bar {x}) \leqslant 
\max_{1 \leqslant l \leqslant d} \varphi\bigl(x(l), x_t(l)\bigr) = \\
=\max_{x \in \Sol(\Phi^*)}\varphi_S(x) = \theta^* .
\end{multline*}
The inequality in this chain follows from the property (\ref{MinMaxProperty}) of 
the set $\Sol'= \argmind_{\bar{x} \in WORK}\varphi(\bar{x})$. The next equality 
is valid due to (\ref{Just}).

Let us prove that the inequality $\varphi(\bar{x}) \ge \theta'$ holds for all 
$\bar{x} \in K^T \setminus \Sol'$.  For labellings $\bar{x} \in WORK \setminus 
\Sol'$ this inequality follows directly from the definition of $\Sol'$. It is 
also true for labellings $\bar{x} \in K^T \setminus WORK$, because
\begin{multline*}
\varphi(\bar{x}) = \max\limits_{i,j \in T}\varphi_{ij}(x_i,x_j) 
\geqslant 
\max_{i,j \in S} \varphi_{ij}(x_i,x_j) = \\
= \varphi_S(x_S) \geqslant \theta^* \geqslant \theta',
\end{multline*}
where $x_S$ is the restriction of $\bar{x}$ to $S$. The second inequality of 
this chain follows from the fact that $x_S \notin \Sol(\Phi^*)$ if $\bar{x} 
\notin WORK$. We obtain, that $\varphi(\bar{x}) \geqslant \theta'$ holds for all 
 $\bar{x} \notin \Sol'$ and  $\varphi(\bar{x}) \leqslant \theta'$ holds for all 
$\bar{x} \in \Sol'$. According to Definition~\ref{MainDefinition}, this means 
that $\Sol' = \argmind_{\bar{ x} \in K^T} \varphi(\bar{x})$. 
\end{IEEEproof} 

It follows from Lemma~\ref{GreedySecurity}, that the Algorithm~\ref{Trivial} is 
applicable for the whole NP-complete class of problems. Its output is either 
{\decline} or a correct solution. Unfortunately, this correct solution is 
obtained only for a very limited set of simple problems. We expand this set by 
replacing the problem $\Phi = \bigl\langle T, (\varphi_{ij} \mid i, j \in T) 
\bigr\rangle$ in step~2 of the algorithm by an equivalent problem $\Omega = 
\bigl\langle T,(\omega_{ij} \mid i,j \in T) \bigr\rangle$. Equivalence means 
here that both problems have the same objective function, i.e.~that  $\max_{i,j 
\in T}\omega_{ij}(x_i,x_j) = \max_{i, j\in T} \varphi_{ij}(x_i,x_j)$ holds for 
any labeling $\bar{x}\in K^T$. Section~\ref{ExclusionOfObject} shows how to 
construct this equivalent problem in such a way, that the extension of 
Algorithm~\ref{Trivial} solves all problems with a majority polymorphism.
\subsection{\label{ExclusionOfObject} Equivalent transformation of a problem. }
\begin{definition}
A structure $\bigl\{\{t, i \} \mid i \in T \setminus \{ t\}  \bigr\}$ defined 
for a set $T$ is called a star with center $t$; a structure $\bigl\{ \{i, 
j\} \mid i, j \in S, i \ne j \bigr\}$ defined for a set $S$ is called a 
simplex.
\end{definition}
The objective function of a problem defined on a star structure $\bigl\{\{t, i 
\} \mid i \in T \setminus \{ t\}  \bigr\}$ is
\begin{equation*}
 \varphi(\bar{x})= \max\limits_{i \in T \setminus \{ t \}} 
 \varphi_{ti}(x_t,x_i) , \hspace{.5em} \bar{x} \in K^T .
\end{equation*}
\begin{definition}
A transformation of a star into a simplex is the transformation of a problem 
$\Phi=\bigl\langle T,(\varphi_{ti} \mid i \in T \setminus \{ t \})\bigr\rangle$,
defined on a star structure, into a problem $\Psi = \bigl\langle 
T \setminus\{t\}, (\psi_{ij}\mid i,j \in T \setminus\{ t\})\bigr\rangle$, 
defined on a simplex structure, such that $\psi_{ij}$ are the projections of 
the objective function of $\Phi$  onto $\{i,j\}$.
\end{definition}
The starting point for the following construction is to represent the objective 
function of the problem $\Phi = \bigl\langle T,(\varphi_{ij} \mid i,j \in T) 
\bigr\rangle$, defined on a simplex, in form of a maximum of two functions
\begin{multline} \label{Development}
 \varphi(\bar{x}) = \max_{i,j \in T} \varphi_{ij}(x_i,x_j) = \\ 
 = \max \bigl(\max_{i\in S} \varphi_{ti}(x_t,x_i), 
 \max_{i,j\in S} \varphi_{ij}(x_i,x_j) \bigr) , \quad \bar{x} \in K^T.
\end{multline}
The first of these functions is the objective function of a problem defined on a 
star. The second one is the objective function of a problem defined on a smaller 
simplex.
\begin{lemma} \label{EquivalencyStatement}
Let $\Phi = \bigl\langle T,(\varphi_{ij} \mid i,j \in T) \bigr\rangle$ be a 
problem, $t \in T$, $S = T \setminus \{t\}$ 
and let $\psi_{ij} \colon K \times K \rightarrow W$ be the projections of the 
function $\max_{l \in S}\varphi_{tl} $ onto $\{i,j\}$. Denote by 
$\omega_{ij}=\max\{\varphi_{ij}, \psi_{ij}\}$ the point-wise maximum 
of the functions $\varphi_{ij}$ and $\psi_{ij}$ for $i,j \in S$. Then the 
equality
\begin{equation} \label{EquivalencyLemmaMy}
 \varphi(\bar{x}) = 
 \max\bigl\{ \max_{i,j \in S}\omega_{ij}(x_i,x_j),
 \max_{i \in S}\varphi_{ti}(x_t,x_i) \bigr\}.
\end{equation} 
holds for any labeling $\bar{x} \in K^T$.
\end{lemma}
\begin{IEEEproof}
The functions $\psi_{ij}$ are the projections of the function $\max_{l \in S}\varphi_{tl} $ 
onto $\{i,j\}$. Therefore, 
the inequality $\psi_{ij}(x_i,x_j) \le \max_{l \in  
S}\varphi_{tl}(x_t,x_l)$ holds for any $\bar{x} \in K^T$ . 
The right-hand side of this inequality does not depend on $(i,j)$, and
$\max_{i,j \in S}\psi_{ij}(x_i,x_j) \le \max_{l \in  
S}\varphi_{tl}(x_t,x_l)$ follows as a consequence. Hence,
\begin{multline*}
  \varphi(\bar{x}) =
  \max\bigl\{ \max_{i,j \in S}\varphi_{ij}(x_i,x_j), 
  \max_{i \in S}\varphi_{ti}(x_t,x_i) \bigr\}= \\
  =\max\bigl\{ \max\limits_{i,j \in S}\psi_{ij}(x_i,x_j), 
  \max\limits_{i,j \in S}\varphi_{ij}(x_i,x_j), 
  \max\limits_{i\in S}\varphi_{ti}(x_t,x_i) \bigr\} ,
\end{multline*}
and (\ref{EquivalencyLemmaMy}) follows immediately.
\end{IEEEproof}
The function on the right-hand side of (\ref{EquivalencyLemmaMy}) can be thought 
of as the objective function of the problem $\Omega = \bigl\langle 
T,(\omega_{ij} \mid i,j \in T) \bigr\rangle$ where 
$\omega_{ij}=\max\{\varphi_{ij}, \psi_{ij}\} $ for $i,j \in S$ and  
$\omega_{tj}= \varphi_{tj} $ for $j \in S$. The problems $\Phi$ and $\Omega$ are 
equivalent because they have the same objective function. Notice that this 
equivalence is not conditioned upon existence of a majority polymorphism. 
The following additional property holds in the event that 
the problem $\Phi$ is invariant under some majority operator.
\begin{lemma} \label{ProjectionStatement}
Let $\Phi = \bigl\langle T,(\varphi_{ij} \mid i,j \in T) \bigr\rangle$ be a 
problem with a majority polymorphism. Let $t \in T$ and $S = T \setminus \{t\}$ 
and let $\psi_{ij} \colon K \times K \rightarrow W$ be the projections of the 
function $\max_{l \in S}\varphi_{tl} $ onto $\{i,j\}$. Then the function 
$\omega: K^S \rightarrow W$, with values defined by 
\begin{equation} \label{ProjectionLemma}
\omega(x_S)=\max\bigl\{ \max_{i,j \in S}\varphi_{ij}(x_i,x_j), 
\max_{i,j \in S}\psi_{ij}(x_i,x_j) \bigr\} ,
\end{equation} 
is the projection of the objective function of $\Phi$  onto $S$.
\end{lemma}
\begin{IEEEproof}
The following chain holds for the projection of the objective function 
$\varphi \colon K^T \rightarrow W$ of $\Phi$ onto the subset $S$ and for any 
labeling $x_S \in K^S$
\begin{align*}
 &\min_{x_t \in K} \varphi(x_S,x_t) = \\
 =&\min_{x_t \in K} \max \bigl\{ \max_{i,j\in S} 
 \varphi_{ij}(x_i,x_j), 
 \max_{i\in S} \varphi_{ti}(x_t,x_i) \bigr\}  = \\
 = &\max \bigl\{ \max_{i,j\in S} \varphi_{ij}(x_i,x_j), 
 \min_{x_t \in K} \max_{i\in S} \varphi_{ti}(x_t,x_i) \bigr\} = \\
 = &\max \bigl\{ \max_{i,j\in S} \varphi_{ij}(x_i,x_j), \max_{i,j\in S} 
 \psi_{ij}(x_i,x_j) \bigr\}.
\end{align*}
The first equality repeats (\ref{Development}), the second equality 
holds because $\max_{i,j\in S} \varphi_{ij}(x_i,x_j)$ does not depend on 
$x_t$ and the third equality holds because the function $\max_{i,j\in S} 
\psi_{ij}$ is the projection of $\max_{i\in S} \varphi_{ti}$ onto $S$, 
as shown in Lemma~\ref{Decomposition}.   
\end{IEEEproof}

In summary, given a problem 
$\Phi = \bigl\langle T,(\varphi_{ij} \mid i,j 
\in T) \bigr\rangle$, 
an arbitrary element $t \in T$ and the subset $S = T \setminus \{t\}$, 
we have constructed the problem
$\Omega = \bigl\langle T,(\omega_{ij} \mid i,j\in T) \bigr\rangle,$
defined by
\begin{align} \label{UpToStarToSimplex}
    \omega_{ij} &=\max\{\varphi_{ij}, \psi_{ij}\}   
    &&\text{for $i,j \in S$,} \\ 
    \omega_{tj} &= \varphi_{tj} &&\text{for $j \in S$.} \nonumber
\end{align}
 According to Lemma~\ref{EquivalencyStatement}, the objective functions of both 
problems are the same, therefore so are their projections onto $S$. Moreover, if 
the problem $\Phi$ has a majority polymorphism then, according to Lemma 
\ref{ProjectionStatement}, the projection of the function $\omega(\bar{x}) = 
\max_{i,j \in T} \omega_{ij}(x_i,x_j)$, $  \bar{x} \in K^T$, onto $S$ is simply 
the function $\omega(x_S) = \max_{i,j \in S} \omega_{ij}(x_i,x_j)$, $x_S \in 
K^S$.

Since the problem $\Omega$ is defined in (\ref{UpToStarToSimplex}) in terms of 
functions $\psi_{ij}$, the procedure of transforming $\Phi$ into $\Omega$ is 
defined up to the procedure of transforming a star into a simplex. This 
transformation can be expressed explicitly due to the following equivalences 
for expressions composed of operations $\max$ and $\min$.

Let $X$ and $Y$ be some finite sets and let $f \colon X \rightarrow W$ and $g 
\colon Y \rightarrow W$ be two functions. Then
\begin{equation} \label{Rule1My}
 \min_{x \in X}\min_{y \in Y} \max \bigl\{ f(x), g(y) \bigr\} = 
 \max \bigl\{ \min_{x \in X}f(x), \min_{y \in Y} g(y) \bigr\} 
\end{equation}
and for any $x \in X$
\begin{equation} \label{Rule2My}
 \min_{y \in Y} \max \bigl\{ f(x), g(y) \bigr\} = 
 \max \bigl\{ f(x), \min_{y \in Y} g(y) \bigr\}.
\end{equation} 
Let $I$ be some finite set, let $X_i$, $i \in I \cup \{ 0 \}$, be finite sets 
and let $f_i:X_i \rightarrow W$, $i \in I \cup \{ 0 \}$ be some functions. Then 
\begin{multline}\label{Rule3My}
    \max \bigl\{ 
    f_0(x_0), \max_{i \in I} \min_{x_i \in X_i}f_i(x_i) 
    \bigr\}= \\
    = \max \bigl\{ 
    f_0(x_0), \max_{i \in I \cup \{ 0 \}} \min_{x_i \in X_i}f_i(x_i)
    \bigr\}
\end{multline}
for any $x_0 \in X_0$.

Using rules (\ref{Rule1My}) - (\ref{Rule3My}), the star-to-simplex 
transformation is constructed in the following way. Let us pick two objects $m,n 
\in S$ and fix them for the following considerations. Denote $R = S \setminus \{ 
m, n \}=T \setminus \{t,m,n\}$. The following chain of equalities holds for the 
values $\psi_{mn}(x_m,x_n)$ of the projection of $\varphi(\bar{x})=\max_{i \in 
S}\varphi_{ti}(x_t,x_i)$ onto $\{m,n\}$:
\begin{align*}
 &\psi_{mn}(x_m, x_n) = \nonumber\\
 &\min_{x_t \in K} \min_{x_R \in K^R} \varphi(x_t,x_m,x_n, x_R) = \nonumber\\
 & \min_{k \in K} \min_{x \in K^R} 
 \max_{i \in S} \varphi_{ti}(k, x_i)= \nonumber\\ 
 &\min_{k \in K} \max 
 \bigl\{ \varphi_{tm}(k, x_m),\; \varphi_{tn}(k, x_n),\;
 \min_{x \in K^R} \max _{i \in R} \varphi_{ti}(k,x_i) \bigr\} = \nonumber\\ 
 &\min_{k \in K} \max 
 \bigl\{  \varphi_{tm}(k, x_m),\; \varphi_{tn}(k, x_n),\;
 \max_{i \in R}\min_{x_i \in K} \varphi_{ti}(k,x_i) \bigr\}= \nonumber\\ 
 &\min_{k \in K} \max \bigl\{
 \varphi_{tm}(k, x_m),\; \varphi_{tn}(k, x_n),\;
 \max _{i \in S}\min_{x_i \in K} \varphi_{ti}(k,x_i) \bigr\}. \nonumber
\end{align*}
The first two equalities are valid by definition. The third one is valid 
according to rule (\ref{Rule2My}), the fourth one is valid according to 
(\ref{Rule1My}) and the fifth one is valid according to (\ref{Rule3My}). The 
following explicit expression for the star-to-simplex transformation
\begin{multline} \label{StarToSimplexExplicitlyMy}
  \psi_{ij}(x_i, x_j) = \\
  = \min_{k \in K} \max \bigl\{
  \varphi_{ti}(k, x_i), \; \varphi_{tj}(k, x_j), \;
  \max _{l \in S}\min_{x_l \in K} \varphi_{tl}(k,x_l)  \bigr\}
\end{multline}
is obtained as a result. 

Algorithm~\ref{ProjMy} implements the transformation of the problem $\Phi$ into 
the problem $\Omega$ based on expressions (\ref{UpToStarToSimplex}) and 
(\ref{StarToSimplexExplicitlyMy}).
\begin{algorithm}{ \bf Equivalent transformation of problems.} \label{ProjMy}\\
{\bf Input:} a problem $\Phi = \Big\langle T, \big(\varphi_{ij}|i, j \in T\big) 
\Big\rangle $ and $t \in T$.\\
{\bf Output:} a problem $\Omega = \Big\langle T, \big(\omega_{ij}|i, j \in 
T\big) \Big\rangle $.\\ 
0. Let $S = T \setminus \{ t \}$;\\
1. for all $k \in K$ compute $q(k) = \max\limits_{i \in S} \min\limits_{x \in 
K}\varphi_{ti}(k,x)$;\\
2. for all $i, j \in S$, $x,y \in K$ compute\\
\SP \SP  $\text{  }  \psi_{ij}(x,y)= 
\min\limits_{k \in K} \max\big\{ \varphi_{ti}(k,x),
\varphi_{tj}(k,y), q(k) \big\}$;\\
\SP \SP $\text{  } \omega_{ij}(x,y) = \max \big\{ \varphi_{ij}(x,y), \psi_{ij}(x,y) 
\big\}$;\\
3. for all $i \in S$, $k, x \in K$ let $\omega_{ti}(k,x) = 
\varphi_{ti}(k,x)$. \qed
\end{algorithm}
The complexity of Algorithm~\ref{ProjMy} is of order $|K|^3 \times |T|^2$.

\subsection {\label{FinalEnd} Solving problems of order two.}
We include Algorithm~\ref{ProjMy} for equivalent transformation into the 
general Algorithm~\ref{Trivial} for exclusion of variables.
\begin{algorithm}{\bf Solving a problem of order two.}\label{Final}  \\
{\bf Input:} a problem $\Phi = \bigl\langle T, (\varphi_{ij}|i, j\in T)
\bigr\rangle $.\\
{\bf Output:} either $\Sol(\Phi)$ or a message \decline. \\
0. If $\abs{T}=2$, $T = \{ a,b \}$ \\
\SP\SP then $\Sol\bigl(\Phi\bigr) =
\argmind\limits_{(x_a, x_b) \in K^2 }
\varphi_{ab}(x_a,x_b)$;\\
else\\
1. pick $t \in T$, let $S = T \setminus \{t\}$;\\
2. using Algorithm \ref{ProjMy} construct the problem \\
\SP\SP    $\Omega = \bigl\langle T, (\omega_{ij} \mid i,j \in T) \bigr\rangle 
$;\\
3. using Algorithm \ref{Final} construct the set \\
\SP\SP    $\Sol\bigl(\Omega^*\bigr)$, $\Omega^* = \bigl\langle S, (\omega_{ij} 
\mid i,j \in S) \bigr\rangle $; \\
4. if at least one labeling $x \in \Sol\bigl(\Omega^*\bigr)$ \\
\SP\SP fulfills the inequality\\
\SP\SP $ \max\limits_{i,j \in S} \omega_{ij}(x_i, x_j) < 
\min\limits_{k \in K } \max\limits_{i \in S} \omega_{ti}(k,x_i)$, \\
\phantom{4. }then return the message \decline;\\
5. construct the auxiliary set \\
\SP\SP $WORK = \bigl\{(x, x_t)  \in K^T \bigm| x \in \Sol(\Omega^*), x_t \in K 
\bigr\}$;\\
6. find 
$\Sol' = \argmind\limits_{\bar{x} \in WORK}
\max\limits_{i,j \in T}\omega_{ij}(x_i,x_j)$. \qed
\end{algorithm}
The complexity $Q(T)$ of Algorithm \ref{Final} is the sum of complexities of 
p.0-6:\\
the complexity of p.0 is of order $\abs{K}^2 + d \times \log\abs{K}$;\\
the complexity of p.1 may not be taken into account;\\
the complexity of p.2 is of order $\abs{T}^2 \times \abs{K}^3$;\\
the complexity of p.3 is $Q(T \setminus \{ t \})$;\\
the complexity of p.4 is of order
$d \times (\abs{T}^2 + \abs{T} \times\abs{K})$;\\
the complexity of p.5 is of order $d \times \abs{K}$;\\
the complexity of p.6 is of order
$d \times \abs{T}^2 \abs{K} + d \times \abs{K} + d \times
\log (d \times \abs{K})$.

Adding up these values and excluding components with order less than that of 
others, we obtain the recursive expression
\begin{equation*}
 Q(T) = Q(T \setminus \{t\}) + {\cal O}(\abs{T}^2 \times \abs{K}^3 
 + d \times \abs{T}^2 \times \abs{K} + d \times \log d)  
\end{equation*}
and the following explicit expression for $Q(T)$
\begin{equation*}
 Q(T)={\cal O}(\abs{T}^3 \times \abs{K}^3 + 
 d \times \abs{T}^3 \times \abs{K} + 
 \abs{T} \times d \times \log d).  
\end{equation*}
The following theorem justifies Algorithm~\ref{Final}. 
\begin{theorem}\label{FinalResult} Let $\Phi = \bigl\langle T, (\varphi_{ij} 
\bigm | i, j\in T) \bigr\rangle$ be an input problem for Algorithm~\ref{Final} 
and $\Sol'$ be a subset of $d$ labellings constructed by the algorithm.
\begin{enumerate}
 \item If the algorithm does not return a \decline\ message then 
\begin{equation} \label{COMMON1}
 \Sol'= \argmind_{\bar{x} \in X^T} \max_{i,j \in T} \varphi_{ij}(x_i,x_j) ,
\end{equation}
what means that the algorithm returns a valid solution of the problem.
\item If the problem $\Phi$ has a majority polymorphism then
(\ref{COMMON1}) is valid without any further condition.
\end{enumerate}
\end{theorem}
\begin{IEEEproof}
Let $\Omega = \bigl\langle T, (\omega_{ij} \mid i,j \in T) \bigr\rangle$ and 
its restriction 
$\Omega^* = \bigl\langle S, (\omega_{ij} \mid i,j \in S) \bigr\rangle$ onto $S$ 
denote the auxiliary problems constructed in steps~2 and 3 of the algorithm.
 
Let us prove the first statement of the theorem. If the algorithm has not 
returned a \decline\ message, then the inequality 
\begin{equation} \label{IfLemmaFinal}
  \max_{i,j \in S} \omega_{ij}(x_i,x_j) \geqslant
  \min_{k \in K } \max_{i \in S} \omega_{ti}(k,x_i)
\end{equation} 
holds for each labeling $x \in \Sol(\Omega^*)$.
It follows from Lemma~\ref{GreedySecurity} 
and condition~(\ref{IfLemmaFinal}) that
\begin{equation*}
 \Sol' = \argmind_{\bar{x} \in X^T} 
 \max_{i,j \in T} \omega_{ij}(x_i,x_j) .
\end{equation*}
By taking into account equivalence of problems $\Phi$ and $\Omega$, which 
follows from Lemma~\ref{EquivalencyStatement}, we obtain (\ref{COMMON1}). 

Let us prove the second statement. Since $\Phi$ is assumed to have a majority 
polymorphism, it follows from Lemma~\ref{ProjectionStatement} that the objective 
function of the problem $\Omega^*$ is the projection of the objective function 
of the problem $\Phi$ onto $S$. This means that inequalities 
(\ref{IfLemmaFinal}) are valid for all $x \in K^S$ including $x \in 
\Sol(\Omega^*)$. Consequently, the proof can be completed by 
repeating from hereon the proof of the first statement.
\end{IEEEproof}
\section{\label{Algorithmproperty} Conclusion.}
We have analyzed the problem of finding $d$ best labellings 
$\bar{x} \colon T\rightarrow K$, where $T$ and $K$ are finite sets and the 
quality $\varphi \colon K^T \rightarrow W$ of a labeling is given in a format 
similar to constraint satisfaction theory. This addresses the search
of $d$ smallest numbers in a set of $\abs{K^T}$ numbers. If the function  
$\varphi$ is invariant under a majority operator, then this problem is reduced 
to a sequence of $(\abs{T}-2)$ essentially easier problems. Each of them 
seeks $d$ smallest numbers in a set of $\abs{K} \times d$ numbers. 
In particular, if $d=1$ then  the $\abs{T}$-variate minimization is reduced to 
$(\abs{T}-2)$ univariate minimizations.

This strength would be severely weakened, if the behavior of the algorithm on 
problems with no majority polymorphism was not known. This would require an 
additional algorithm for testing the existence of a majority polymorphism 
for the input problem. We do not know such an algorithm and expect it to be 
quite complex. The advantage of the proposed algorithm is that it does not 
require such control. It copes with the whole NP-complete class of minimax 
problems of a certain format. For any such problem the algorithm returns either 
the solution or a \decline\ message. The latter is possible only if the problem 
has no majority polymorphism.
\bibliographystyle{plain}
\bibliography{biblio}{}

\end{document}